\documentclass[lettersize,journal,compsoc]{IEEEtran}
\usepackage{amsmath,amsfonts}
\usepackage{algorithmic}
\usepackage{algorithm}
\usepackage{array}
\usepackage{caption}
\usepackage[caption=false,font=normalsize,labelfont=sf,textfont=sf]{subfig}
\usepackage{textcomp}
\usepackage{stfloats}
\usepackage{url}
\usepackage{ragged2e}
\usepackage{verbatim}
\usepackage{graphicx}
\usepackage{multirow} 
\usepackage{cite}
\usepackage{color}
\begin{document}

\title{Infinite Motion: Extended Motion Generation via Long Text Instructions}

\author{Mengtian~Li, Chengshuo~Zhai, Shengxiang~Yao, Zhifeng~Xie$^{*}$, Keyu~Chen$^{*}$, Yu-Gang~Jiang, \IEEEmembership{Fellow, IEEE}

\thanks{\textbullet\ $^{*}$: Corresponding author}
\thanks{\textbullet\  Mengtian~Li is with Shanghai University, Fudan University. E-mail: mtli@$\left\{shu \setminus  fudan \right\}$.edu.cn}
\thanks{\textbullet\  Chengshuo~Zhai, Shengxiang~Yao and Zhifeng~Xie are with Shanghai University.  E-mail:$\left\{zcshuo \setminus yaosx033 \setminus zhifeng \_ xie \right\}$ @shu.edu.cn}
\thanks{\textbullet\ Keyu~Chen is with Tavus Inc.. E-mail: keyu@tavus.dev.}
\thanks{\textbullet\ Yu-Gang~Jiang is with the School
of Computer Science, Fudan University. E-mail: ygj@fudan.edu.cn}
}

\markboth{Journal of \LaTeX\ Class Files,~Vol.~14, No.~8, August~2021}%
{Shell \MakeLowercase{\textit{et al.}}: A Sample Article Using IEEEtran.cls for IEEE Journals}

\IEEEpubid{0000--0000/00\$00.00~\copyright~2021 IEEE}

\IEEEtitleabstractindextext{
\setcounter{figure}{0}
\captionsetup{type=figure}
\noindent\includegraphics[width=0.92\textwidth]{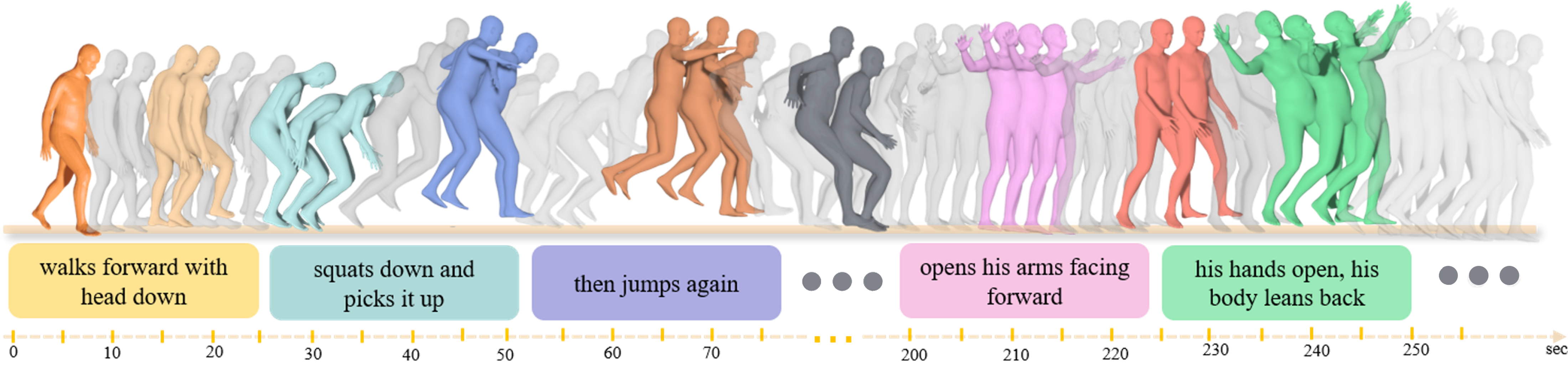}
\captionof{figure}{{\bfseries Infinite motion}: We propose a novel method for generating infinite motions, based on the timestamps featured in our HumanML3D-Extend dataset. This approach not only enables the generation of extremely long motions but also facilitates precise control over actions within specific time intervals.}
\label{fig:1}

\vspace{1em}
\begin{abstract}
\justifying
In the realm of motion generation, the creation of long-duration, high-quality motion sequences remains a significant challenge. This paper presents our groundbreaking work on "Infinite Motion", a novel approach that leverages long text to extended motion generation, effectively bridging the gap between short and long-duration motion synthesis. Our core insight is the strategic extension and reassembly of existing high-quality text-motion datasets, which has led to the creation of a novel benchmark dataset to facilitate the training of models for extended motion sequences. A key innovation of our model is its ability to accept arbitrary lengths of text as input, enabling the generation of motion sequences tailored to specific narratives or scenarios. Furthermore, we incorporate the timestamp design for text which allows precise editing of local segments within the generated sequences, offering unparalleled control and flexibility in motion synthesis. 
We further demonstrate the versatility and practical utility of "Infinite Motion" through three specific applications: natural language interactive editing, motion sequence editing within long sequences and splicing of independent motion sequences. Each application highlights the adaptability of our approach and broadens the spectrum of possibilities for research and development in motion generation. Through extensive experiments, we demonstrate the superior performance of our model in generating long sequence motions compared to existing methods.

\end{abstract}

}
\maketitle

\section{Introduction}
\label{sec:introduction}
\IEEEPARstart{T}{he} generation of dynamic and realistic motion sequences is a cornerstone of modern computer graphics, with applications spanning from interactive media to cinematic storytelling. While there has been a notable improvement in the quality and realism of short-duration text-driven motion generation, the synthesis of long-duration motions remains an elusive goal. This is primarily due to two interrelated challenges: the high cost and difficulty associated with the collection of long human motion sequence data, and the limitations of current motion generation models in handling long textual inputs.

In this work, we aim to address the above issues for long-duration motion generation from the root cause, the lack of a high-quality benchmark dataset with long-duration motion data and text description. We take a thorough investigation of existing text-motion datasets and propose an efficient idea to extend the short-duration motion sequences as well as the text labels to meet the length requirement for long-term motion synthesis. This dataset is not only a testament to the potential of long text in driving motion synthesis but also serves as a critical resource for the development and evaluation of future models in this domain.

The creation of this benchmark dataset represents a significant milestone for several reasons. Firstly, it provides a rich and diverse collection of long-sequence motions that can be used to train and refine motion generation models. By incorporating a wide range of realistic motion sequences and corresponding text descriptions, the dataset ensures that models are exposed to the complexity and variability required to generate long-duration motions that are both coherent and engaging. The dataset is analyzed on multiple dimensions e.g. frame numbers, text length, and motion diversity, and all data points indicate that it covers a wider range of motion distribution than the existing ones. Secondly, the dataset is designed with the unique challenges of long sequence generation in mind. STMC~\cite{STMC} uses some text prompts in the dataset as the base "atomic" actions. Similarly, we have also applied new preprocessing to the textual information in our dataset.
It includes detailed annotations and temporal metadata that allow models to better understand and learn the nuances of extended motion sequences. This level of detail is crucial for addressing the long-duration motion generation problem and ensuring that generated motions maintain coherence over longer time spans. Moreover, our dataset is complemented by an innovative approach to handling long textual inputs. By introducing a timestamp mechanism, we are able to process and generate motions based on arbitrary lengths of text. This not only enhances the dataset quality because of contextually rich motions but also provides a level of precision that allows for localized editing within the sequences.

Beyond the efforts of building a novel benchmark dataset, we propose a subsequent diffusion-based motion generation model that is capable of handling long text input and generating faithful motion sequences. As a known issue, diffusion models~\cite{motiondiffuse,petrovich2021action,mofusion,mld,MDM} generally struggle with memorizing and generating long-duration data and hence it's also non-trivial to extend the motion generation~\cite{shafir2023human,teach,diffcollage} capability from short to long. In this work, we propose a novel idea to fine-tune the motion diffusion models with an extended plug-and-play module called \textit{timestamp stitcher}. Based on the timestamp design, we successfully maintain high-quality motion generation for short-duration data while innovatively enabling arbitrarily long text input. Specifically, we make the \textit{timestamp stitcher} capable of decomposing long texts into multiple short pieces with timestamps, then we adopt a fine-tuned motion generation model to predict the motion sequences for each piece, finally, we train a temporal conditional diffusion network to stitch all motion pieces together with a smooth transition. 

Through extensive experimentation, we demonstrate the effectiveness of our benchmark dataset, named HumanML3D-Extend, in training models that can generate high-quality, long-duration motion sequences. The results showcase a significant leap forward in the capability of motion generation models, setting a new standard for the synthesis of extended motions.

\section{relwork}

\subsection{Text to Motion Datasets}
HumanML3D~\cite{Guo_2022_CVPR_humanml3d} is currently the largest 3D human motion dataset with text descriptions and is the most commonly used dataset in text-driven human motion generation tasks.
The human body format data consists of a standard 22 joint points
The dataset contains 14,616 human motion sequences and 44,970 text descriptions. The entire text description consists of 5371 different words for a total duration of 28.59 hours.
The average motion sequence length was 7.1 seconds, while the average description length was 12 words. The initial motion sequences in this dataset are from AMASS~\cite{mahmood2019amass} and HumanAct12~\cite{guo2020action2motion}.
The motion is scaled to 20 FPS, and those over 10 seconds will be randomly cropped to 10 seconds, then repositioned to the default human skeleton template and rotated correctly to the original Z + orientation.
Kit~\cite{plappert2016kit} is the same author as the HumanML3D dataset and is the original text2motion dataset. 
The human body format in this dataset is represented by 21 joint points, containing 3,911 human motion sequences and 6,278 text annotations. The total vocabulary is 1623. The sample size is very limited compared to HumanML3D.

The Multi-track Timeline (MTT)~\cite{STMC} dataset is designed for controlling motion generation across multiple tracks. It consists of 500 multi-track timelines, each including three prompts distributed across two tracks. Each timeline is automatically generated using a set of 60 manually collected texts representing different "atomic" actions (e.g., “punch with the right hand”, “jump forward”), with annotations for the relevant body parts (e.g., $\sharp$ right arm, $\sharp$ spine). The primary aim is to control various body parts on different tracks within the same timeframe, enhancing fine-grained control over human motion sequences.

However, none of their data is enough to train current long-duration motion generation tasks. HumanML3D-Extend dataset contains 35,000 motion data and 105,000 different text descriptions. Our dataset is the first long-duration motion sequence dataset, and its number of motions and the richness of text descriptions meet the requirements of long motion training.

\subsection{Human Motion Generation}
At present, the research on human motion generation tasks has received extensive attention. Action category(~\cite{guo2020action2motion, petrovich2021action,teach}) , music(~\cite{tseng2023edge,zhuang2022music2dance,mofusion,lee2019dancing,mu},) or unconditional (~\cite{un,un1l,un2l,un3,un4}). 

Since text is the simplest and easiest way to communicate between humans and computers, text-driven has been the focus of research.
Most of them are enhancements to the generated human motion sequences.
Similar to T2M~\cite{Guo_2022_CVPR_humanml3d} and TEMOS~\cite{temos}, text and motion sequences are mapped into a discrete space, where they are subsequently matched with each other.
MotionDiffuse~\cite{motiondiffuse} applies the diffusion model(~\cite{diffu, diffu1}) to motion generation, 
MDM~\cite{MDM} uses the diffusion model of the transformer to continuously predict the original sample sequence.
TM2T~\cite{guo2022tm2t} and T2M-GPT~\cite{T2MGPT} encodes actions into the codebook and then indexes the text into the action codebook.
Mofusion~\cite{mofusion} proposes a new diffusion model generation framework on which some additional physical loss of human action can be added.
There are also some other derivative tasks. MotionGPT~\cite{jiang2024motiongpt} can handle a variety of motion tasks, such as motion2text and text2motion, motion2motion,i.e. MLD~\cite{mld} combines VAE and diffusion model to perform generation tasks in the latent space and improves the generation speed on the premise of ensuring the quality of the motion.
MoMask~\cite{guo2024momask} proposes a  masked modeling framework that employs a hierarchical quantization scheme and bidirectional Transformers, enhancing the fidelity and detail of the generated motions.
ReMoDiffuse~\cite{zhang2023remodiffuse} introduces a diffusion-model-based motion generation framework that integrates a retrieval mechanism to refine the denoising process, thereby enhancing the generalizability and diversity of text-driven motion generation.
LAVIMO~\cite{LAV} introduces a three-modality learning framework that integrates human-centric videos to bridge the gap between text and motion, utilizing a specially designed attetntion mechanism to enhance alignment and synergy among text, video, and motion modalities.
"Motion patches"~\cite{yu2024exploring} employ transfer learning to enhance cross-modal latent space modeling between human motion and text description, aiming to address data scarcity.
However, the above methods primarily focus on the generation of short sequence motions, which have limited practical utility.

For the Long human motion generation task. When Teach~\cite{teach} generates processing for the next section of a motion, additional frames of the previous motion are introduced, and text information is added to guide the generation of the next motion. MultiAct~\cite{MultiAct} uses motion to generate intermediate segments and the next pose sequence, which are then stitched together to grow the sequence. DiffCollage~\cite{MultiAct} combines graph models with generative models, stitching them together with graphs and then removing duplicates to stitch. PriorMDM~\cite{shafir2023human} combines two MDM~\cite{MDM} modules and adds splicing modules between the two modules for long sequence splicing. STMC~\cite{STMC} proposes a timeline control method for text-driven motion synthesis, which allows for the specification of multiple overlapping text prompts on a temporal timeline. STMC~\cite{STMC} primarily focuses on controlling the body in a multi-track space. However, Current long-sequence generation tasks are unable to achieve precise control over specific time segments. 
Our work incorporates timestamp information into the long text descriptions in the HumanML3D-Extend dataset. We designed a model to segment and process the text based on timestamps to generate corresponding short motion sequences. The~\textit{timestamp stitcher} then seamlessly splices these short motion sequences together. This approach enables the generation of infinitely long motion sequences with precise time control and allows for the re-editing of specific segments within the long motion sequences.

\begin{figure*}
\centering
\includegraphics[width=1\textwidth]{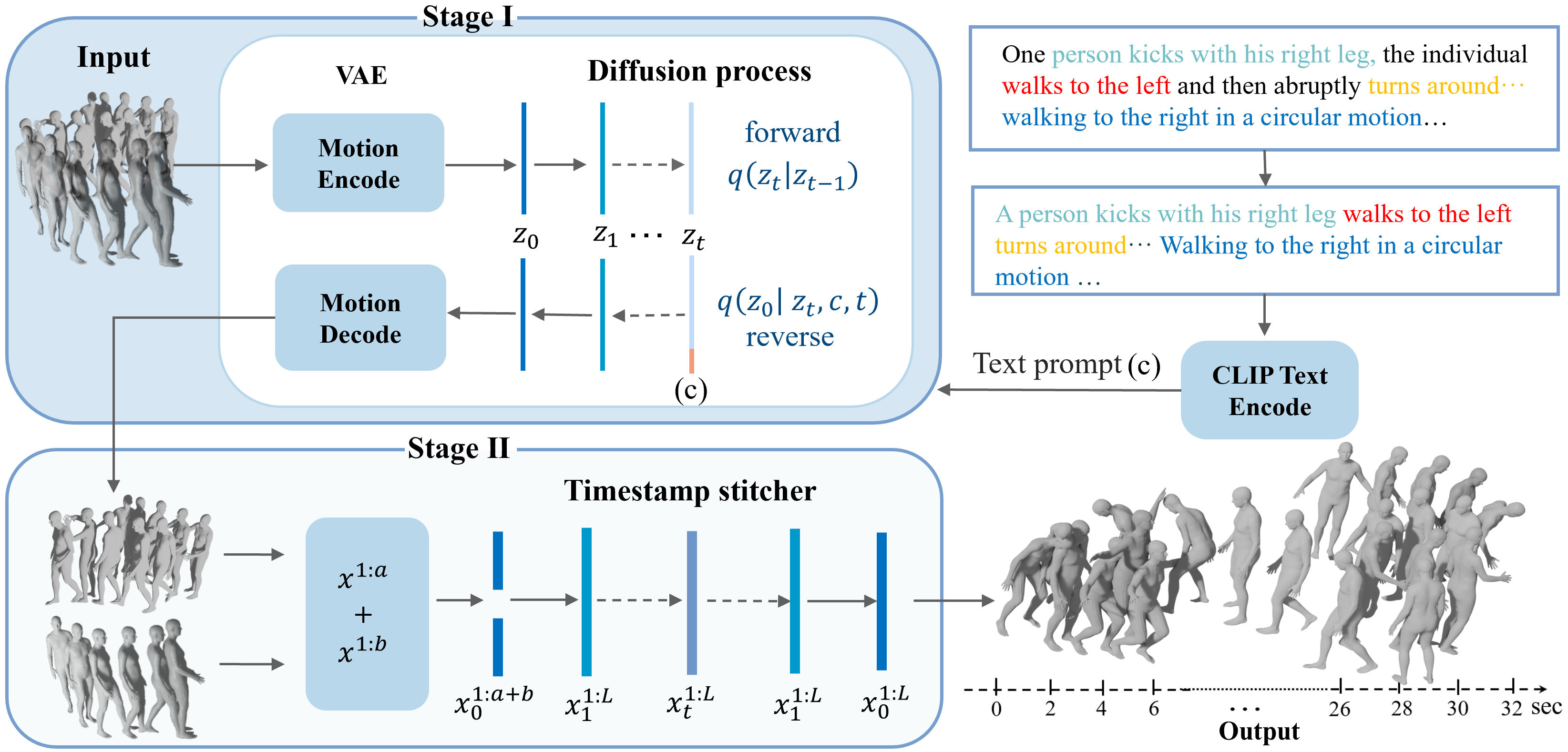}
\caption{Infinite Motion Pipeline: Our model consists of two stages. In Stage I, a diffusion process occurs in the latent space, simultaneously generating multiple segments of short motion sequences. In Stage II, the \textit{timestamp stitcher} concatenates these short motion segments to form an infinite sequence of motions.}
\label{fig:pipeline}
\end{figure*}

\section{method}
\label{sec:method}
In this section, we begin by briefly outlining the workflow of our baseline approach in Sec.~\ref{subsec:Pre}.
Due to the current lack of extensive datasets for long-duration text-to-motion generation task, we create a novel benchmark dataset \textit{HumanML3D-Extend} containing long texts and lengthy motions in Sec.~\ref{dataset}.
Furthermore, we propose a fine-tuned strategy on the baseline method MLD~\cite{mld} and incorporate a novel extended motion generation idea called \textit{timestamp-stitcher}. In Sec.~\ref{model}, we describe the training pipeline of our model \textit{Infinite Motion} and demonstrate the additional timestamp editing feature enabled by our method.

\begin{figure*}
\centering
\includegraphics[width=\textwidth]{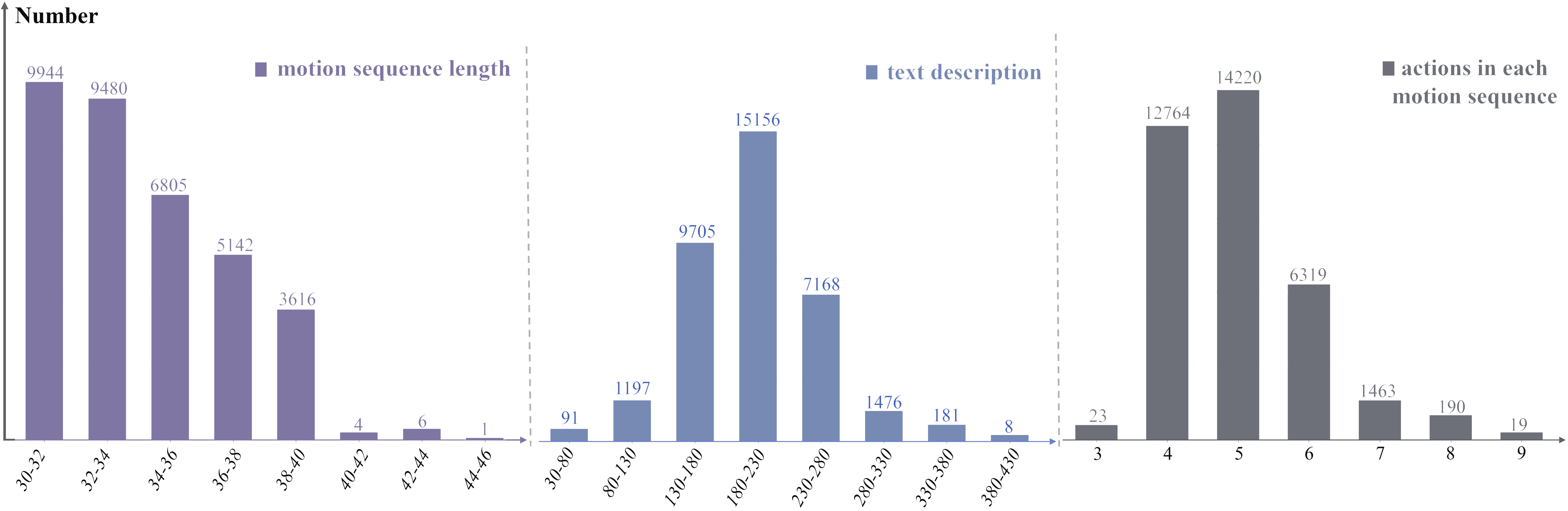}
\caption{The data distribution of the HumanML3D-Extend dataset. {\bfseries Left}: The horizontal axis represents the number of frames in a motion sequence. This chart shows the distribution of frame counts across various motion sequences.
{\bfseries Middle}: The horizontal axis represents the number of words in each text description. This chart shows the distribution of word counts in text descriptions. {\bfseries Right}: The horizontal axis represents the number of actions in each motion sequence. This chart shows the distribution of action counts in each motion sequence.}
\label{fig:dataset1}
\end{figure*}

\subsection{Preliminary}
\label{subsec:Pre}
We utilize a Variational Autoencoder (VAE)~\cite{ling2020character} as a generative model to learn the underlying representation of input motion data. 
As shown in stage one in Figure~\ref{fig:pipeline}, the motion encoder maps the sequence of human motion $x^{L}$ with arbitrary length $L$. into the latent space $z_{0}$. The motion decoder takes the latent code $z_{0}$ as input and reconstructs it back to the original motion data space, generating $\hat{x} ^{L}$ similar to the input motion data $x^{L}$.

In the low-dimensional latent space, we implement a diffusion based model from MLD~\cite{mld} which is capable of adding noise on the latent code $z_ {0}$ with corresponding text feature input and denoising reversely. During the forward process, the diffusion network gradually loads the noise along time $t$, and predicts the noise adding process $z_{0}\longrightarrow z_{t}$. The forward process can be formulated as follows, where $\alpha _{t}$ is a hyper parameter for controlling the noise, $z_{0}$ represents the initial motion and $z_{t}$ denotes the noise at time step.$z_{t}$ represents the noisy sequence in the latent space at time $t$.

\begin{equation}
q(z_{t}|z_{t-1}) = \mathcal{N} (\sqrt{\alpha _{t} }z_{t-1},(1-\alpha)I).
\end{equation}

Then the reverse process of the diffusion model is performed according to the input text conditions $c$, and the $z_{t}$ is gradually denoised to the original $z_{0}$ code following the formulation below, $\epsilon_\theta$ represents the denoiser:
\begin{equation}
Z_{t-1}=\epsilon _\theta (Z_{t},t,c).
\end{equation}

The denoising process necessitates a series of iterations to successfully generate noise-free samples, which is time-consuming. However, by conducting this process within the latent space, we can significantly accelerate the diffusion learning process, thereby reducing the required time and computational resources. This streamlined approach not only optimizes the workflow but also enhances the overall effectiveness and reliability of the denoising process.
\subsection{HumanML3D-Extend Dataset}
\label{dataset}

We introduce the HumanML3D-Extend dataset, an expansion of the HumanML3D~\cite{Guo_2022_CVPR_humanml3d} motion data. HumanML3D-Extend comprises 35,000 motion data entries and 105,000 distinct text descriptions, with each motion entry corresponding to three text descriptions. The motion sequences range from 600 to 935 frames in length, recorded at 20 frames per second, resulting in a total duration of 330.16 hours. Individual motion sequences last between 30 and 46 seconds. The average and median lengths of the text descriptions are 64.75 and 63 words, respectively. This dataset is the first to offer long motion sequences, providing a sufficient quantity of motions and rich text descriptions to support the training of models for long motion generation.

Table~\ref{tab} presents a comparison of our HumanML3D-Extend dataset with two existing motion-text datasets, HumanML3D~\cite{Guo_2022_CVPR_humanml3d} and KIT Motion-Language~\cite{plappert2016kit}. The table details differences in dataset duration, text length, and motion diversity, demonstrating that our dataset covers a broader range of motion distribution compared to the existing datasets. To the best of our knowledge, the HumanML3D-Extend dataset represents the largest and most diverse collection of scripted human motions to date.

\setlength{\tabcolsep}{6pt}
\renewcommand{\arraystretch}{1.5}
\begin{table}[H]
    \centering 
    \caption{Comparisons of 3D human motion-language datasets.} 
    \label{tab}
    \resizebox{1\linewidth}{!} 
	{ 
        \begin{tabular}{{c| c| c|c |c| c }}
            \hline
             Dataset&\textbf{$\sharp$Motions} & \textbf{$\sharp$texts}  &\textbf{Duration} &\textbf{max duration} &\textbf{min duration}\\  \hline
            KIT-ML& 3,911 &6,278 &10.33h & -&- \\ 
            HumanML3D & 14,616& 44,970 & 28.59h &10s&2s \\
              \textbf{HumanML3D-Extend} & 35,000& 105,000 & 330,16h & 46s &30s\\\hline
        \end{tabular}
        }
\end{table}
The following describes the production process of our dataset in two steps, data collection and refinement.

\noindent {\bfseries Data collection.} Initially, we refined the original dataset by filtering out sequences with fewer than 10 frames, removing all short sequences. Subsequently, we analyzed the starting and ending poses within all sequences, filtering based on the principle that the facing directions of the two frames should be consistent. This ensures that the poses of the frames before and after stitching do not differ too greatly, avoiding unnatural results in the stitched sequences.
Then, we aligned the root position of the subsequent motion with the root position of the previous motion. Finally, we selected the last 5 frames of the previous motion and the first 5 frames of the next motion for interpolation to achieve smooth intermediate transitions.

The detailed algorithm steps are described in Algo~\ref{alg:data_collection}. Where $B(t)$ represents the point on the curve at parameter $t$, $P_{0}$ is the starting point, $P _{1}$ is the control point, and $P _{2}$ is the ending point. The parameter $t$ ranges from 0 to 1. 

\begin{algorithm} 
    \caption{Data collection process} 
    \label{alg:data_collection} 
    \begin{algorithmic}
        \STATE \textbf{Input:} Some short human motion sequences;
        \STATE \textbf{Output:} Long human motion sequences greater than 600 frames;
        \STATE \textbf{Steps:}
         \item 1. Delete data smaller than 10 frames,
         \begin{center}
                {$len(motion) \ge 10$} : delete .
        \end{center}
        \item 2. Calculate the head $(x_{h},y_{h},z_{h})$ and neck $(x_{n},y_{n},z_{n})$ direction vectors ($\overrightarrow{D}$) and  unit vector ($\hat{D}$),
        \begin{center}
        \centering  $\overrightarrow{D} = (x_{h},y_{h},z_{h}) - (x_{n},y_{n},z_{n})$\vspace{0.2cm} ,
          \STATE \centering $\hat{D} = \frac{\overrightarrow{D}}{\left \| \overrightarrow{D}  \right \| }$,\vspace{0.2cm}
         
          \STATE {$\hat{D}_{start} $ == $ \hat{D}_{end}$} : splicing,
         \STATE {$\hat{D}_{start} $  !=  $\hat{D}_{end}$} : select the next motion .
        \end{center} 
         \hspace{0.5cm} \item 3. Align the root position of the subsequent motion with the root position of the previous motion,
        \begin{center}
         \centering $ offset = joint_{s} - joint_{e}$,
         \STATE \centering $ newjoint_{e} = joint_{e} + offset$.
        \end{center} 
         \hspace{0.5cm} \item 4. Select the last 5 frames of the previous motion and the first 5 frames of the next motion for interpolation,
        \begin{center}
        \centering torch.cat( $joint_{s} \left [ -5:,: \right ]$ , $newjoint_{e} \left [ :5,: \right ] $.)$\longrightarrow I\left [ 10,: \right ]$.
        \end{center} 
         \hspace{0.5cm} \item 5. Interpolate($I$) these frames($i = 10$) to achieve smooth transitions,
        \begin{center}
         \centering $I\left [ i,: \right ] = \left (1-\frac{i}{10-1}   \right ) \cdot I_{s} + \frac{i}{10-1}\cdot I_{e} $.
        \end{center} 
         \hspace{0.5cm} \item 6. Quadratic Bézier curve calculation for feet,
        \begin{center}
         \centering $B(t) = (1-t)^{2}\cdot P _{0} +2(1-t)t\cdot P_{1} + t^{2} \cdot P_{2}$.
        \end{center} 
        \hspace{0.5cm} \item 7. Keep sequences longer than 600 frames,
        \begin{center}
                  {$len(motion)\ge 600$} : save,
                \STATE {$len(motion) < 600$} : continue. 
                \end{center} 
                
        \STATE \textbf{End}
    \end{algorithmic} 
\end{algorithm}

\begin{figure}
\centering
\includegraphics[width=0.5\textwidth]{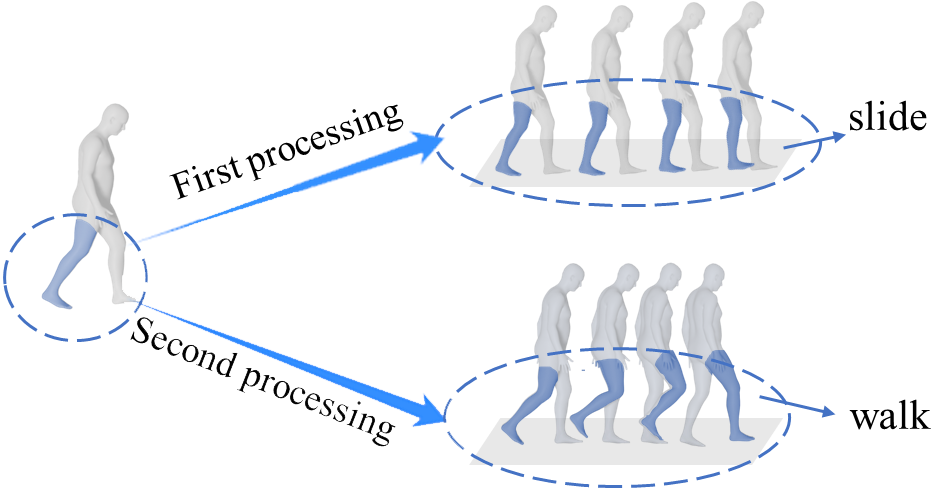}
\caption{Solve foot sliding issue: After the initial processing, there is a sliding issue with the foot position. Perform quadratic Bézier curve processing on the foot position to ensure it conforms to a normal human walking posture.}
\label{fig:foot}
\end{figure}

\noindent {\bfseries Data refinement.} The motion data processed solely through smooth interpolation techniques may cause sliding of the foot position, which does not conform to normal human movements. To address this issue, a method employing quadratic Bézier curve interpolation for splicing foot positions was adopted, as illustrated in Figure~\ref{fig:foot}. This method entails localizing the position of the feet within the sequence. To ensure standard human foot movements, the interpolation process is applied sequentially: initially, one foot is processed independently while maintaining the other in a static position. Subsequently, the other foot is processed in the later frames of the sequence.

\begin{figure}
\centering
\includegraphics[width=0.5\textwidth]{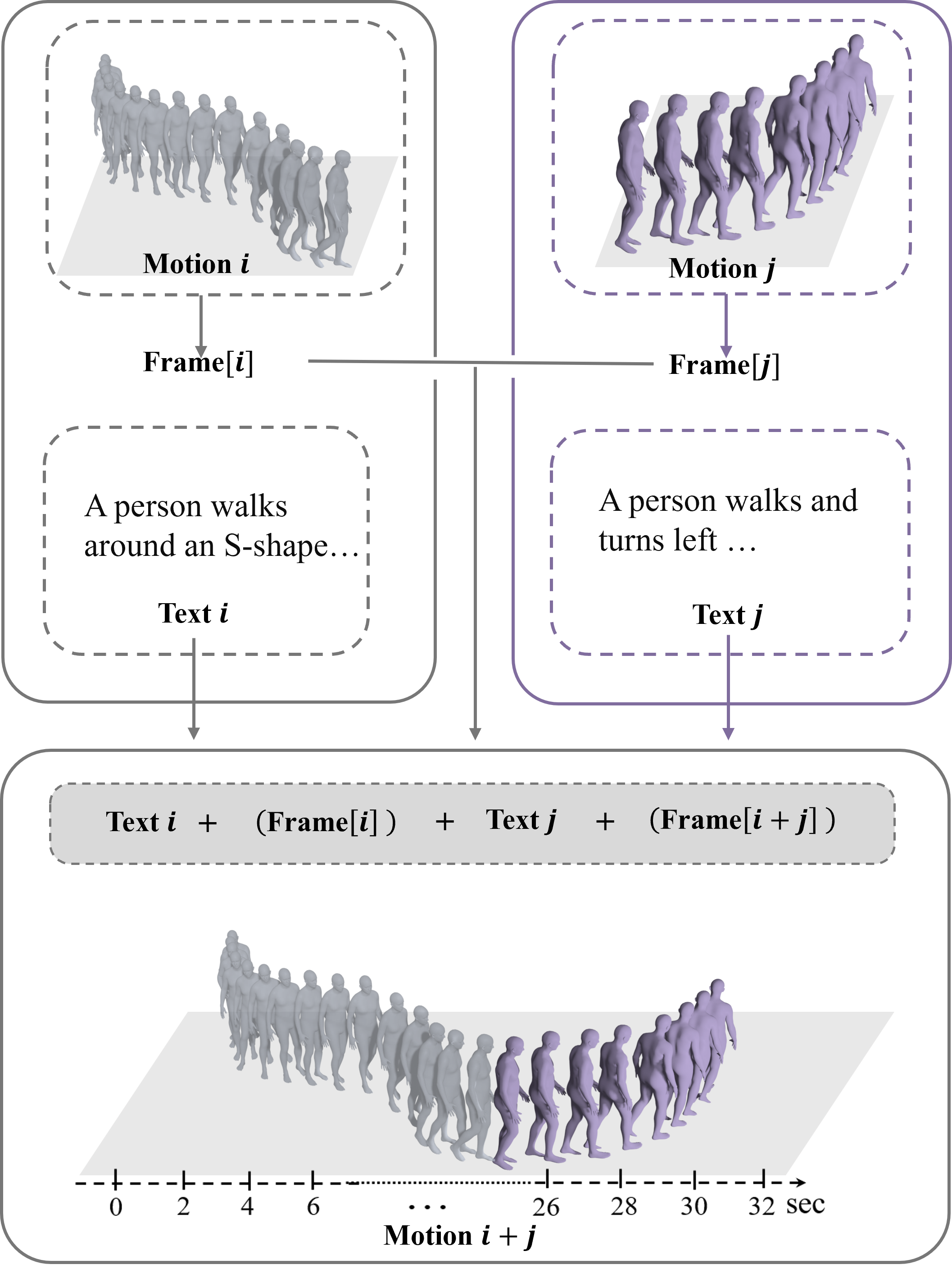}
\caption{The process of timestamp insertion: The sequence lengths (frame[$i$], frame[$j$]) and corresponding text descriptions (text[$i$], text[$j$]) of two motion sequences are extracted. The sequence length of the preceding motion (frame[$i$]) is inserted as a timestamp at the junction of the two text descriptions. The subsequent timestamp is the sum of the lengths of the two motion sequences (frame[$i + j$]).}
\label{fig:timestamp}
\end{figure}

As shown in Figure~\ref{fig:timestamp}. To enhance the model's ability to recognize lengthy text description and ensure consistency between the generated motions and the text. We add timestamps to the text descriptions. During the motion stitching process, we extract and preserve the duration information for each short action. This duration information is incorporated into text messages to ensure precise matching between the text descriptions and the long human motion sequences. This allows the model to accurately align the motions with the corresponding text during training, thereby enhancing the accuracy of the generated outputs.

\begin{figure}
\centering
\includegraphics[width=0.5\textwidth]{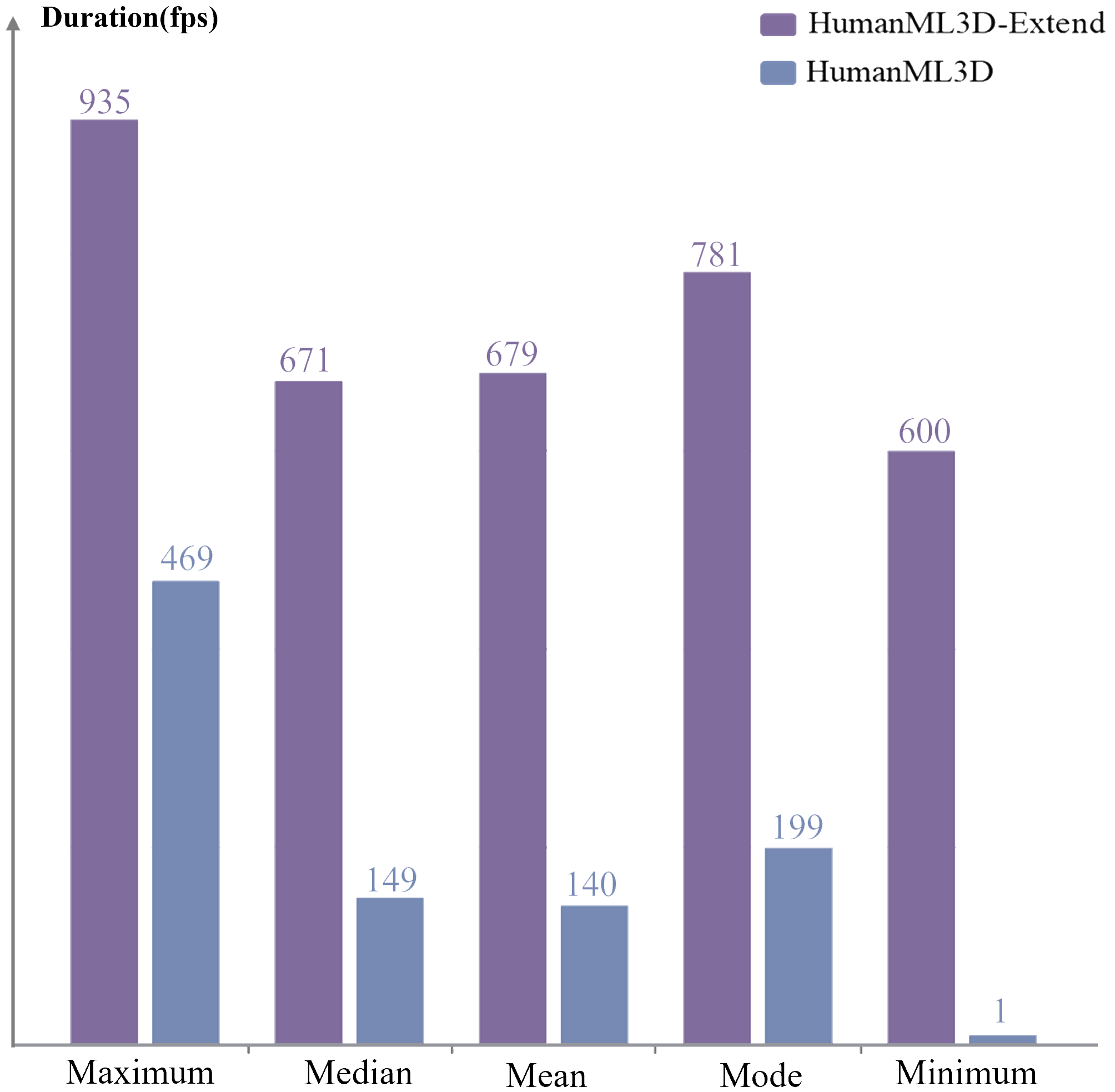}
\caption{Comparison with humanML3D dataset: the maximum duration, median duration, mean duration, mode duration and minimum duration are compared respectively. The vertical axis represents the number of frames.}
\label{fig:data1}
\end{figure}
\noindent {\bfseries Dataset analysis.} We conduct an analysis on our dataset along multiple dimensions. In Figure~\ref{fig:dataset1}, we illustrate the distribution of quantities concerning motion sequence length, text descriptions, and action counts in each sequence.
It turns out that the distribution of quantities is relatively concentrated in every respect. The duration of motion sequences is generally maintained between 30 and 40 seconds. Most of the text descriptors for a motion sequence range between 130 and 280 words. Most motion sequences contain 4 to 5 actions, with a few containing 7 to 9 actions.

We have demonstrated the richness of the HumanML3D-Extend dataset across multiple dimensions. Subsequently, we conduct a detailed comparison between the HumanML3D-Extend dateset and the widely utilized HumanML3D dataset, focusing on the duration aspects of the data. We assessed several key statistical metrics to comprehensively understand these temporal characteristics of both datasets. These metrics included the maximum duration, minimum duration, median duration, modal duration (the most frequently occurring duration), and the average duration.  Figure~\ref{fig:data1} shows that the HumanML3D-Extend dataset not only surpasses the HumanML3D dataset in terms of longer motion sequences but also demonstrates a more uniform distribution of durations.

\begin{figure}
\centering
\includegraphics[width=0.5\textwidth]{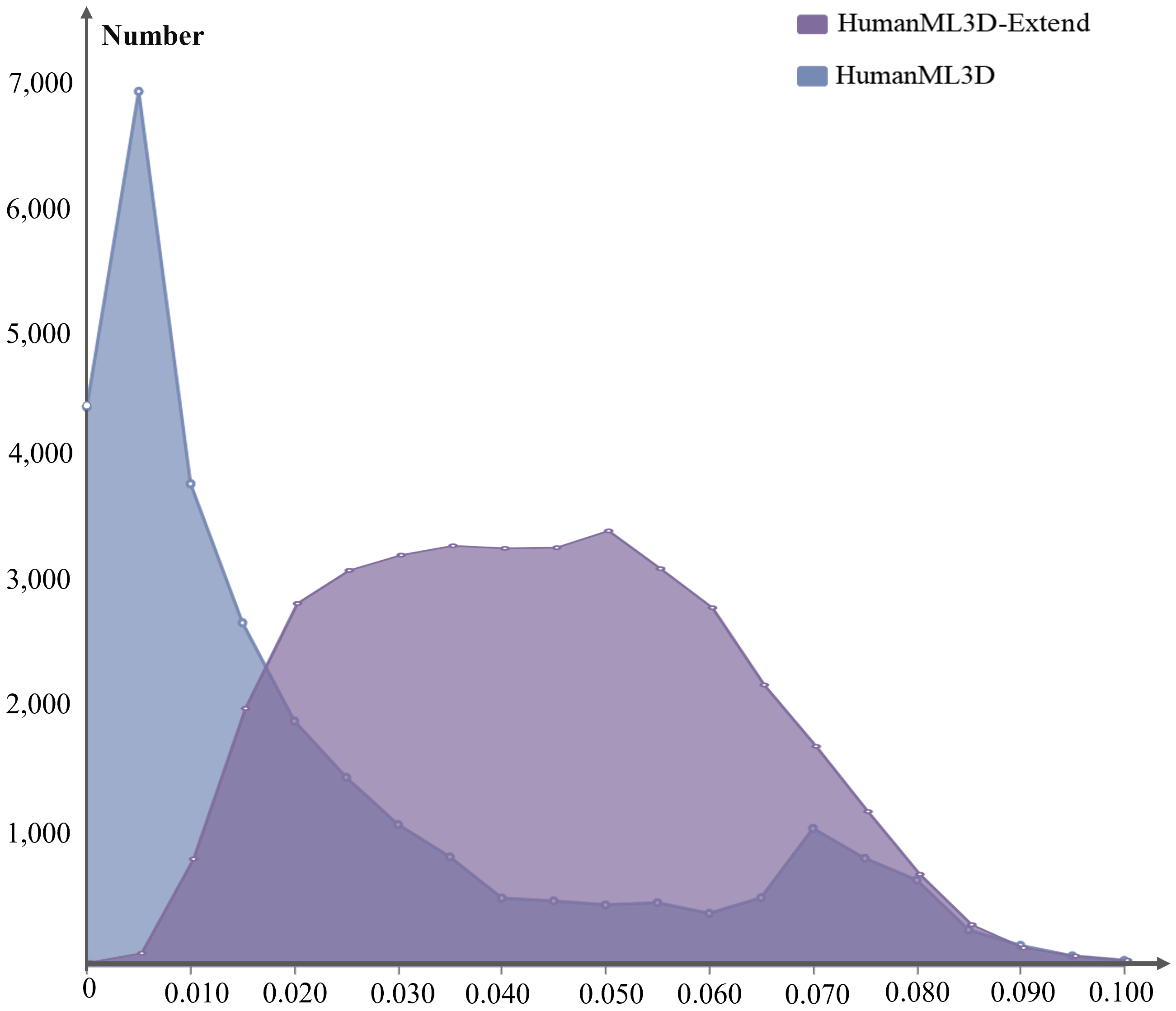}
\caption{The variance distribution of two datasets: the horizontal axis represents the range of variance, and the vertical axis indicates the quantity.}
\label{fig:variance}
\end{figure}

\begin{figure*}
\centering
\includegraphics[width=\textwidth]{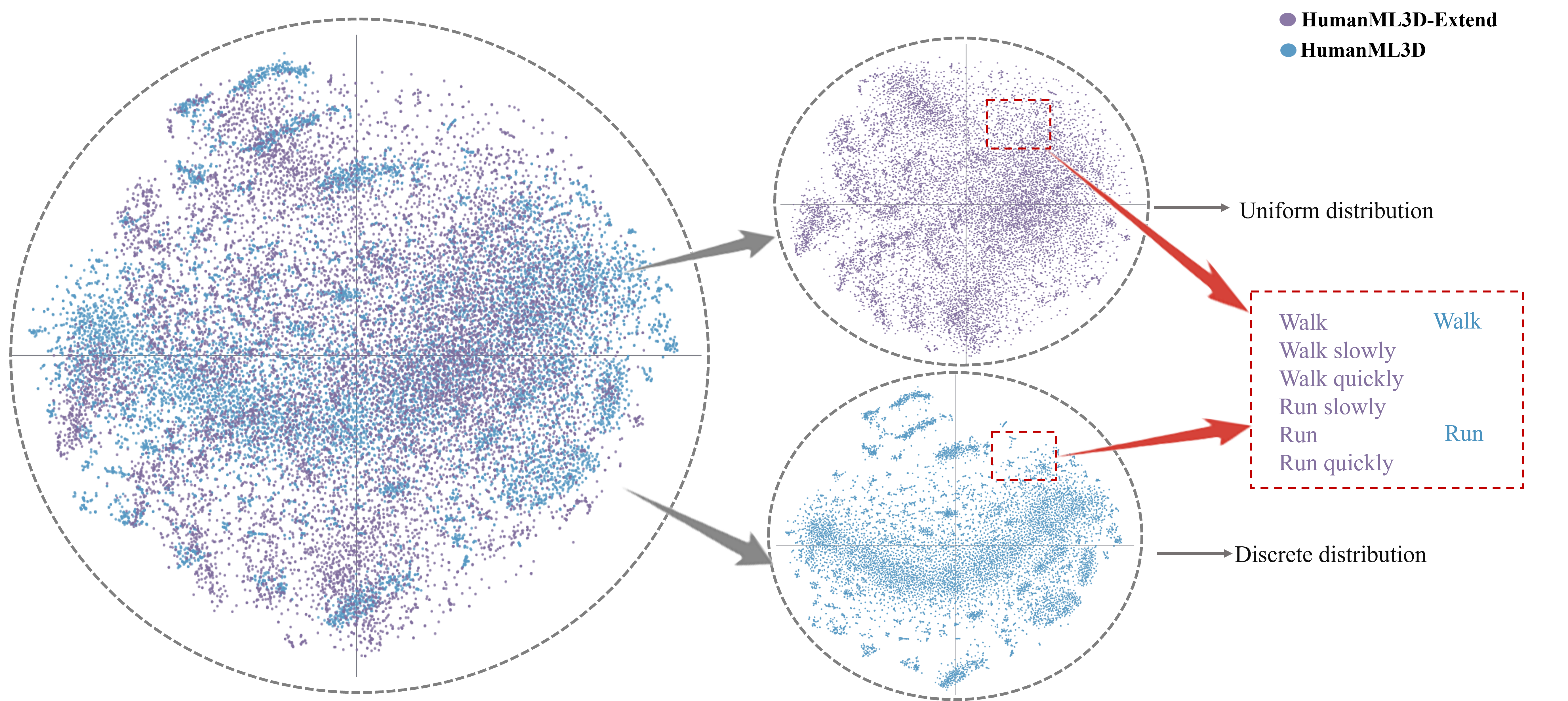}
\caption{We performed random sampling on the HumanML3D dataset and the HumanML3D-Extend dataset to ensure that the sampled motion sequences have the same number and frames. We then visualized the embedding distributions of these sequences. The distribution ranges of the two datasets are comparable, but the HumanML3D-Extend dataset exhibits a more uniform distribution compared to the HumanML3D dataset.}
\label{fig:dataset3}
\end{figure*}

We computed the variance of all motion data in the HumanML3D and HumanML3D-Extend datasets to compare the diversity of actions within individual motion sequences. As shown in Figure~\ref{fig:variance}, the variance distribution in the HumanML3D-Extend dataset is more uniform and concentrated. Additionally, we visualized the embedding distributions of motion sequences from both datasets, randomly selecting an equal number of 200-frame sequences, as shown in Figure~\ref{fig:dataset3}. Although the range of embedding distributions is comparable between the two datasets, the HumanML3D dataset shows more dispersion. The two results demonstrate that the HumanML3D-Extend dataset not only contains more diverse actions within individual sequences but also exhibits a very uniform overall data distribution.

\begin{figure}
\centering
\includegraphics[width=0.5\textwidth]{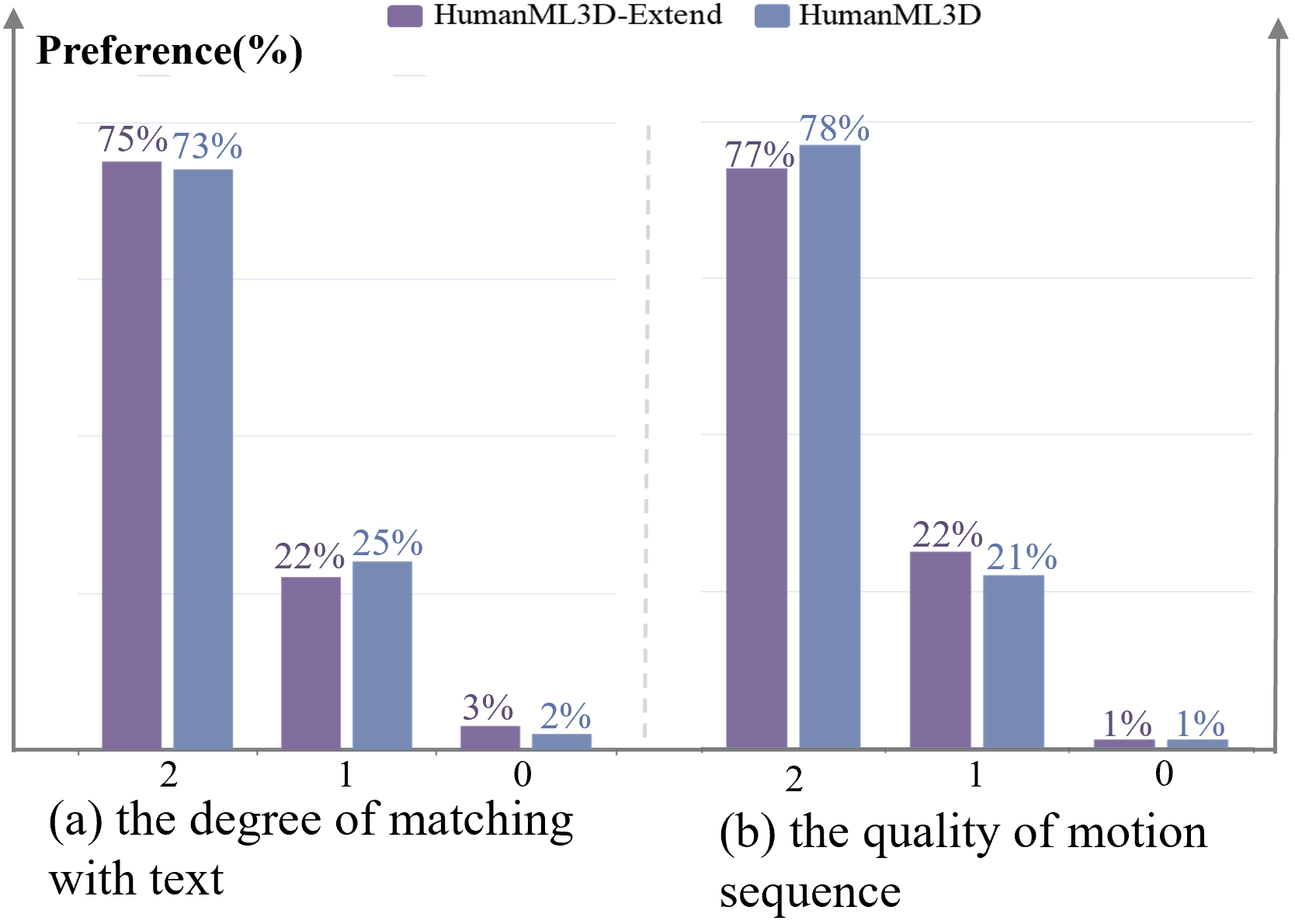}
\caption{User Study: The rating results for the two datasets in terms of (a) the matching degree between the motion and the text description, and (b) the quality of the human motion sequence where 2 indicates excellent performance and 0 indicates average performance. }
\label{fig:data4}
\end{figure}

We conducted a user study on two datasets. We randomly selected 100 samples from both the HumanML3D and HumanML3D-Extend datasets, and had participants rate them on a scale of -2 to 2. For each sample, we identified the most frequently occurring score.  As shown in Figure~\ref{fig:data4}, the results indicated that none of the samples received a score of -1 or -2. The HumanML3D-Extend dataset was found to be comparable to the HumanML3D dataset in terms of text-to-motion alignment and the quality of motion. Additionally, the HumanML3D-Extend dataset demonstrated significantly longer motion durations and a richer variety of actions per sample, making it more suitable for long sequence generation tasks.

\subsection{Infinite Motion}
\label{model}
Infinite Motion is primarily divided into two stages. In the first stage, long text information is processed, and high-quality short motion sequences are generated using our fine-tuned baseline model on the HumanML3D-Extend dataset. In the second stage, the~\textit{timestamp stitcher} can seamlessly stitch together two short motion sequences. After these two stages, Infinite Motion can understand long text inputs and generate infinite and natural human motion sequences.

\noindent {\bfseries Baseline model.}  In the task of text-driven long-motion generation, we choose to use MLD~\cite{mld} as the pre-training model and fine-tune it with our dataset. MLD~\cite{mld} maps motion data to a latent space via VAE and performs diffusion tasks in this latent space, significantly reducing computational overhead in both the training and inference stages. However, since MLD~\cite{mld} is only used for recognizing short texts and generating short motion sequences, it cannot effectively recognize entire long text information. To solve this problem, we utilize the timestamp feature in the HumML3D-Exend dataset to segment long text and accurately match the action data in long motion sequences. This enables efficient recognition of the meaning of long text information and achieves frame-level control over human motion sequences.

When handling lengthy textual information, the model identifies and aligns the text with corresponding frame numbers $C, L \longrightarrow \left \{ (c_{1},l_{a}), (c_{2},l_{b}) \cdots(c_{n},l_{m}) \right \}$. Where $C$ represents a long text description and $L$ represents the length of the total motion sequence. $c_{1},c_{2},\cdots c_{n}$ correspond to shorter text descriptions derived from the long text $C$. $a,b,\cdots m$ represent motion fragments of varying lengths. 
During the inference stage, multiple short motion sequences undergo diffusion processes simultaneously in the latent space. By utilizing cross-attention mechanisms, different textual information is integrated into the reverse diffusion process, thereby generating multiple short motion segments that precisely correspond to the given text.

\begin{figure}
\centering
\includegraphics[width=0.5\textwidth]{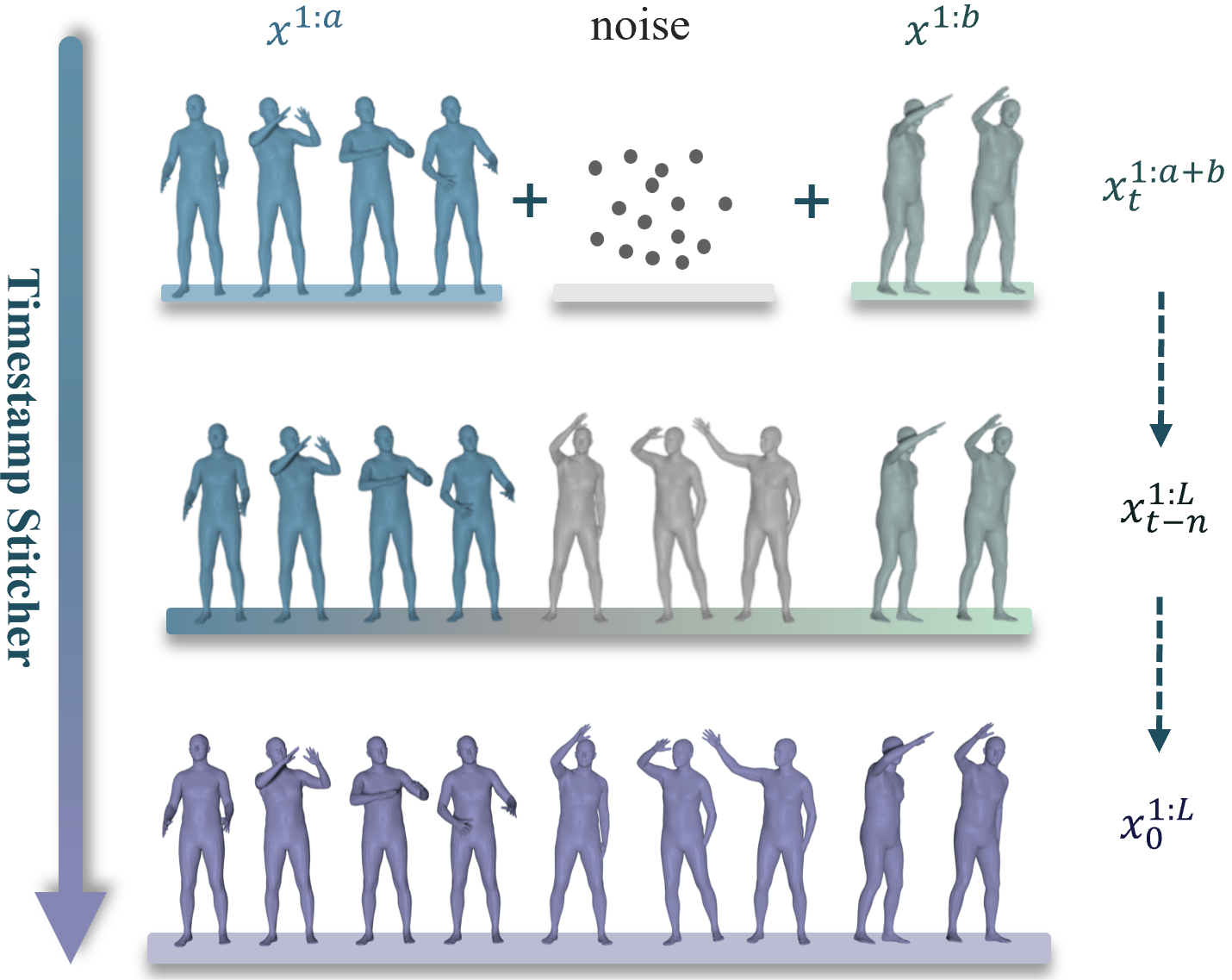}
\caption{~\textit{timestamp stitcher}: The module takes in two segments of motion sequences $x^{1:a}$ of arbitrary length $a$, and $x^{1:b}$ of arbitrary length $b$, introduces random noise between them, and then undergoes the reverse process of the diffusion model. $L$ represents the total length after splicing, $x_{t}$ represents the noisy sequence at time $t$. After from $t$ to $t_{0}$, this gradually combines the two separate motion sequences into one complete and natural motion sequence $x_{0}^{1:L}$.}
\label{fig:4}
\end{figure}

\noindent {\bfseries~\textit{Timestamp stitcher}.}  Our~\textit{timestamp stitcher} is trained based on the diffusion model. During training, a segment of human motion sequence is input, and 10$\%$ of the frames in the middle of the sequence are randomly selected and masked. Noise is added only to the masked portion, and then the entire sequence is fed into the denoising process for denoising. In addition,~\textit{timestamp stitcher}$(TS)$ is learned with classifier-free diffusion guidance~\cite{ho2022classifier}. it learns both the conditioned and the unconditioned distribution with 10$\%$ dropout~\cite{saharia2022photorealistic} of the samples, the formula is as follows:
\begin{equation}
TS_{s} (x_{t} ,t,c) = TS_{s} (x_{t} ,t,\emptyset ) + s\cdot ((TS_{s} (x_{t} ,t,c)-TS_{s} (x_{t} ,t,\emptyset ))
\end{equation}
where $x_{t}$ represents the noisy sequence at time $t$. During the inference stage, the model splices together the motion segments generated in the previous stage with different frame numbers, Initially, noise is introduced to connect them, forming a noisy segment $x_{t}$. Subsequently, through a step-by-step denoising process, formulated as follows:
\begin{equation}
q(x_{t-1}|x_{t}) = \mathcal{N}(x_{t-1} ;\mu_{\theta}  (x_{t},t),\beta _{t}I)
\end{equation}
$\mu_{\theta}$ and $\beta _{t}$ are both hyperparameters.
the noise is gradually removed. As shown in the Figure~\ref{fig:4}, restoring the noisy segment to an intermediate frame sequence that represents a natural transition between the two motions. 

\section{experiments}
In this section, we first introduce the experiments setup  in Sec.~\ref{sub:set}. Then, we compared the proposed method with other short sequence generation tasks. Test them on HumanML3D test-set in Sec.~\ref{sub:com1} and on HumanML3D-Extend test-set in Sec.~\ref{sub:com2}. At last, in Sec.~\ref{sub:comlong}, we demonstrate the splicing capability of the ~\textit{timestamp stitcher} in two aspects:  by using metrics specific to the splicing process, and by comparing it with other long-sequence task.

\subsection{Setup}
\label{sub:set}
{\bfseries Implementation details. }
We first fine-tune the MLD~\cite{mld} on HumanML3D-Extend dataset in stage-one, and then train the~\textit{timestamp stitcher} in stage-two. During training, the diffusion steps are set to 1K.
In the fine-tuning phase, about 600 epochs are trained, and the batch size is set to 128. It took about three days. Training the {\textit{timestamp stitcher} 600 epochs took four days. They all use the NVIDIA 4090 GPU.

\noindent {\bfseries Evaluation Metrics.} 
For fair comparisons, we adopt the following evaluation metrics: (1) Frechet Inception Distance (FID) evaluates the overall quality of motion by measuring the distribution differences between high-level features of generated motions and real motions. A lower FID indicates a closer resemblance to real motions. (2) R-Precision and Multimodal Distance metrics assess the semantic alignment between the input text and the generated actions. R-Precision measures the degree to which the generated actions correspond to the given text descriptions, while Multimodal Distance evaluates the consistency between multiple generated actions and the text. (3) Diversity measures the variability and richness of the generated action sequences. (4) The Multimodality metric evaluates the diversity of motions generated from the same text input.

For the transition frames of the {\textit{Timestamp stitcher}, due to the different tasks, We adopt the metrics mentioned in TEACH~\cite{teach}, specifically Average Positional Error (APE) and Average Variational Error (AVE), as described in (~\cite{ghosh2021synthesis,temos}). These metrics are measured separately on the root joint and the rest of the body joints. "Mean local" and "Mean global" refer to the joint positions in the local (with respect to the root) or global coordinate systems, respectively.
We also measured the transition distance, defined as the Euclidean distance between the poses of two consecutive frames around the transition time. Similar to STMC\cite{STMC}, we compute this distance in the local coordinate system of the body to more effectively capture the transitions of individual body parts. This metric is highly sensitive to abrupt changes in posture. If the transitions between motions are natural, the transition distance should not be excessively high.

\subsection{Comparison on HumanML3D Benchmark Test-set}
\label{sub:com1}

\noindent We perform quantitative comparisons with  MotionDiffuse~\cite{motiondiffuse}, Tm2T~\cite{t2m}, MDM~\cite{MDM}, MLD~\cite{mld}, ReModiffuse~\cite{zhang2023remodiffuse} and MoMask~\cite{guo2024momask} at the same text-to-motion task setting. However, they are all trained on the HumanML3D dataset. To ensure the fairness of the test and the training validity,  we used HumanML3D-Extend train-set for training and used the HumanML3D test-set for testing with other methods.
\begin{table}
    \centering 
    \caption{Quantitative results on the HumanML3D Test-set. {\underline{Underlined}} indicate the best results, {\textbf{bold}} indicate the second-best results. An upward arrow (↑) indicates that a higher value is better, a downward arrow (↓) indicates that a lower value is better, and a rightward arrow (→) indicates that the value closest to the ground truth is the best.} 
    \label{tab1}
    \resizebox{1\linewidth}{!} 
	{ 
        \begin{tabular}{{c c c c c c c c}}
            \hline
             &\textbf{R precision} & \textbf{R precision}   &\textbf{R precision} & \multirow{2}{*}{\textbf{FID↓}}&\multirow{2}{*}{\textbf{MM Dist↓}}&\multirow{2}{*}{\textbf{Diversity→}} &\multirow{2}{*}{\textbf{MModality↑}} \\ 
            & top 1↑&top 2↑&top 3↑& \\ \hline
            GroundTrurth~\hfill & 0.51 & 0.70 & 0.79& 0.08&2.97&9.50&-\\ 
            MotionDiffuse~\hfill\cite{motiondiffuse} & 0.48& 0.65& \textbf{0.78}& 1.63&\underline{2.93}&\underline{9.40}& - \\ 
            Tm2T~\hfill\cite{t2m} & 0.42& 0.61& 0.72& 1.57 &3.47&8.62&2.42\\      
            MDM~\hfill\cite{MDM} & 0.41& 0.60& 0.70& 0.63 &3.65&\underline{9.40}&\underline{2.85}\\ 
            MLD~\hfill\cite{mld} &0.46 &0.65 &0.75 & 0.41 &3.28 &9.80&2.60\\
            ReModiffuse~\hfill\cite{zhang2023remodiffuse} &\textbf{0.48}&\textbf{0.67} &0.77 &\textbf{0.12} &3.04 &9.22 &2.49\\
            MoMask~\hfill\cite{guo2024momask}&\underline{0.52}&\underline{0.71}&\underline{0.80}&\underline{0.04}&\textbf{2.95}&9.64&1.23\\
            
             \textbf{Ours} &0.45 &0.64 &0.75 & 0.90 &3.33&9.17& \textbf{2.64} \\ \hline
        \end{tabular}
    } 
\end{table}

\begin{table}
    \centering 
    \caption{Quantitative results on the HumanML3D-Extend Test-set.} 
    \label{tab2}
    \resizebox{1\linewidth}{!} 
	{ 
        \begin{tabular}{{c c c c c c c c}}
            \hline
             &\textbf{R precision} & \textbf{R precision}&\textbf{R precision} & \multirow{2}{*}{\textbf{FID↓}} &\multirow{2}{*}{\textbf{MM Dist↓}} &\multirow{2}{*}{\textbf{Diversity→}} &\multirow{2}{*}{\textbf{MModality↑}} \\ 
            & top 1↑&top 2↑&top 3↑& \\ \hline
           GroundTrurth~\hfill& 0.29  & 0.45 & 0.55& - &3.43&6.31& -\\ 
           Tm2T~\hfill\cite{guo2022tm2t} & 0.13& 0.23& 0.31 &6.53&\textbf{5.04}&\textbf{6.64}&3.12 \\   
        MotionDiffuse~\hfill\cite{motiondiffuse} &0.14 &\underline{0.27}& \underline{0.37}&3.62&\underline{4.55}&7.17 & - \\ 
            MDM~\hfill\cite{MDM} &0.03 &0.06 &0.09& 45.75&8.48&4.18&3.7 \\
            MoMask~\hfill\cite{guo2024momask}&0.11&0.19&0.27&4.68&5.27&7.12&2.03\\
             \textbf{Ours} &\underline{0.16}& \underline{0.27}&\textbf{0.35} &\underline{2.28}& 5.73&\underline{6.16}&\underline{7.91}\\ \hline
        \end{tabular}
    }
\end{table}

The training data of the HumanML3D-Extend is much longer than HumanML3D dataset, which increases the challenge of model training to a certain extent. However, Table~\ref{tab1} shows that the models trained on the HumanML3D-Extend dataset and tested on the HumanML3D dataset test-set perform comparably to other models trained on the HumanML3D dataset. This indicates that although our model is trained on a long motion sequence dataset, it is also effective for generating short motion sequences.

\subsection{Comparison on HumanML3D-Extend Benchmark Test-set}
\label{sub:com2}

To demonstrate the effectiveness of HumanML3D-Extend benchmark for long motion generation tasks, we also tested other methods on the HumanML3D-Extend test-set. As shown in Table \ref{tab2}, It can be observed that our model generally outperforms other models significantly. Notably, our baseline model MLD~\cite{mld}, due to the limitations of VAE, cannot accept motion sequences longer than 500 frames, and ReModifusse~\cite{zhang2023remodiffuse} preprocessed the HumanML3D dataset, resulting in both models being unable to be tested on the HumanML3D-Extend test-set.

Experimental results demonstrate that, compared to existing methods, our model exhibits enhanced accuracy in generates motion sequences of superior quality. Although MotionDiffuse~\cite{motiondiffuse} slightly outperforms our model on R-Precision and MM-Dist, it is important to note that MotionDiffuse has a very long testing time, requiring about 71 hours to sample 1000 instances. Our HumanML3D-Extend test-set contains 5000 samples, making it computationally expensive and time-consuming to test MotionDiffuse on the entire set. Therefore, the results for MotionDiffuse are based on testing 1000 samples, not the complete test-set.

\subsection{Comparison of~\textit{Timestamp Stitcher}'s Capability.}
\label{sub:comlong}
To validate the effectiveness of our~\textit{timestamp stitcher}, we evaluated the splicing results of the~\textit{timestamp stitcher} using metrics from TEACH~\cite{teach}. Table~\ref{tab3} presents the quantitative results for the~\textit{timestamp stitcher} generating 5-frame, 10-frame, and 15-frame sequences. we found that the best results were achieved when stitching 5 frames. 
It should be emphasized that since our Infinite Motion model just relies on the~\textit{timestamp stitcher} for transient transitions between motions, we do not need to generate transition sequences with excessive frame numbers.

\begin{table*}
\centering
 \caption{Quantitative results with different frame. "TS" denotes the \textit{timestamp stitcher}. \underline{Underlined} indicate the best results.} 
    \label{tab3}
    \resizebox{1\linewidth}{!} 
    {
\begin{tabular}{c c c c c c c c c c }
\hline
& \multirow{2}{*}{\textbf{Transition Dist.↓}} & \multicolumn{4}{|c}{\textbf{Average Positional Error↓}} & \multicolumn{4}{|c}{\textbf{Average Variance Error↓}} \\  

& & \multicolumn{1}{|c}{root joint} & global traj. & mean local & mean global & \multicolumn{1}{|c}{root joint} & global traj. & mean local & mean global \\ \hline
"TS" (5 frames) & 3.924 &  \multicolumn{1}{|c}{\underline{0.208}} & \underline{0.135} & \underline{0.112} & \underline{0.216} &  \multicolumn{1}{|c}{\underline{0.125}} & \underline{0.122} & \underline{0.016} & \underline{0.135}\\
"TS" (10 frames) & 3.440 &  \multicolumn{1}{|c}{0.256} & 0.188 & 0.118& 0.270&  \multicolumn{1}{|c}{0.195} & 0.190 & 0.024 & 0.210 \\
"TS" (15 frames) & \underline{2.173} &  \multicolumn{1}{|c}{0.266} & 0.202 & 0.119 & 0.283 &  \multicolumn{1}{|c}{0.201} & 0.196 & 0.027 & 0.218 \\\hline
\end{tabular}
}
\end{table*}

\begin{table}
    \centering 
    \caption{Quantitative results compared to DoubleTake. "TS" denotes the \textit{timestamp stitcher}. "DT" denotes the DoubleTake. \underline{Underlined} indicate the best results.} 
    \label{tab5}
    \resizebox{1\linewidth}{!} 
	{ 
        \begin{tabular}{{c  c c c c}}
            \hline
             &\textbf{R precision} & \multirow{2}{*}{\textbf{FID↓}} &\multirow{2}{*}{\textbf{MM Dist↓}} &\textbf{Diversity→} \\ 
           &top 3↑& &&(GT = 9.74)\\ \hline
           "DT"(5 frame)~\hfill\cite{shafir2023human} & 0.27 &1.93 &8.63 &8.92 \\
            \textbf{"TS"(5 frame)} & \underline{0.77}&\underline{0.74} &\underline{3.28} &8.75 \\ \hline
             "DT"(10 frame)~\hfill\cite{shafir2023human} & 0.29& 1.46& 8.36& \underline{9.02}\\
             \textbf{"TS"(10 frame)} & 0.75 & 1.06& 3.29& 8.59\\ \hline
        \end{tabular}
    }
\end{table}

 To validate the quality of our~\textit{timestamp stitcher}, we compared~\textit{timestamp stitcher} with DoubleTake, which is specifically designed for splicing tasks, in PriorMDM~\cite{shafir2023human}. For the validity of the experimental results, we trained the~\textit{timestamp stitcher} using the same training dataset as PriorMDM, the HumanML3D dataset. We evaluated the transitions using 5-frame and 10-frame splicing. Table~\ref{tab5} shows the quantitative results on the HumanML3D dataset compared to PriorMDM. The results in Tables~\ref{tab3} and ~\ref{tab5} demonstrate that our~\textit{timestamp stitcher} not only generates smooth splicing sequences but also ensures high-quality splicing sequences.

\begin{figure*}
\centering
\includegraphics[width=\textwidth]{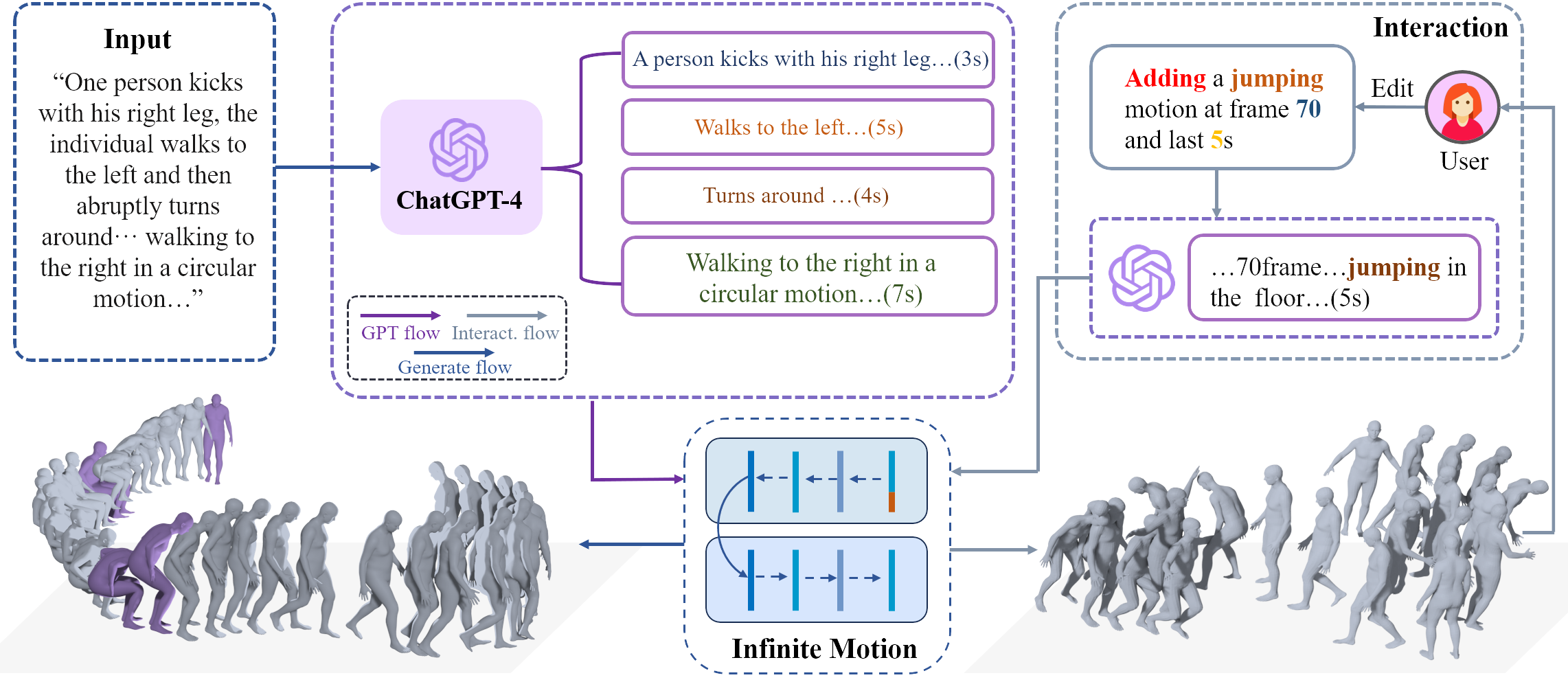}
\caption{The flow of interactive language editing: First, connect to GPT-4. The user inputs a complex language description related to actions, and through GPT-4, we can obtain simple action descriptions and the corresponding frame number requirements. Using Infinite Motion, a complete human motion sequence can be generated. The user can then re-edit the sequence based on the results, allowing for modifications to specific segments of the sequence.}
\label{fig:flow}
\end{figure*}
\section{application}
In this section, we introduce three applications for Infinite Motion. Sec.~\ref{subsec:talk} demonstrates how Infinite Motion can communicate with users seamlessly, fully understanding their requirements for motion sequences and providing accurate feedback. Then, Sec.~\ref{subsec:Precision timestamp editing} introduces the precise control of timestamps, enabling frame-level editing of motion sequences. Sec.~\ref{subsec:Plug and play splicing module} showcases the flexible plug-and-play functionality of the \textit{timestamp stitcher}, which can independently and flexibly perform the splicing of motion sequences.

\subsection{Interactive language editing}
\label{subsec:talk}
Infinite Motion provides users with a simplified editing interface. Initially, it allows users to directly edit motion sequences using natural language commands, thus offering a user-friendly interface for motion editing. As shown in the figure~\ref{fig:flow}, by integrating with advanced language models such as GPT-4, Infinite Motion can better understand the user's requirements for motion, and generate long human motion sequences that closely align with user requests. In addition, Infinite Motion allows users to not only generate human motion sequences based on textual descriptions, but also modify and refine these sequences as needed. Users can flexibly customize motion sequences according to their own requirements, improving the creativity of motion sequences.

Overall, integrating large language models like GPT-4 can enhance Infinite Motion's understanding of user language, which facilitates the generation of longer, more coherent human motion sequences. Utilizing GPT-4's advanced natural language processing capabilities, Infinite Motion is able to accurately interpret the nuances of user instructions and produce motion sequences that closely align with their desires. These capabilities render our model highly practical and user-friendly. Users can easily guide the model using natural language, ensuring smooth and intuitive motion editing and generation.

\subsection{Motion sequence segment editing}
\label{subsec:Precision timestamp editing}
Precision timestamp editing allows users to precisely edit motion sequences within specific time periods of a long motion sequence. This capability is crucial for generating long motion sequences, as it ensures that the desired outcomes are achieved and the quality of the motion is enhanced. As shown in the figure~\ref{fig:app2}, users are required to provide both the frame number range and a text description for the editing process. The frame number range refers to the position that needs to be changed in the original long motion sequence. The text description is the description requirement for the new action. The model then generates a new action sequence based on these textual requirements, which is seamlessly integrated into the original motion sequence at the specified frame numbers. 

This feature has many advantages for generating long motion sequences. Firstly, users can personalize long motion sequences according to their needs. By inputting a range of frame numbers and a text description, they can point out which parts need to be modified and which new actions are required, thus creating a unique sequence of motions that meets their needs. It can also improve efficiency. In the task of generating motion sequences, precise timestamp editing can improve productivity. Instead of completely regenerating the sequence, users can selectively modify specific parts, making the workflow more efficient. Finally, the quality of long motion sequences can be improved. Users can better control the timing and movement of motion, thus improving the quality and fluidity of long motion sequences. This helps to eliminate unnatural parts of the sequence, making the resulting sequence more natural and satisfying.

\subsection{Splicing of independent motion sequences}
\label{subsec:Plug and play splicing module}
Our~\textit{timestamp stitcher} is a plug-and-play stitching module. The term 'plug-and-play' implies that users can input any two human motion sequences, and the~\textit{timestamp stitcher} will generate a transitional sequence that seamlessly integrates the two motions. 
This capability is effective whether transitioning between distinct short human motion sequences or different stages within the same long motion sequence, consistently achieving high-quality results. 
Furthermore, the~\textit{timestamp stitcher} offers  control over the transition sequences through textual inputs. For example, by inputting a descriptor such as 'walking', the~\textit{timestamp stitcher} can interpret this information and generate smooth transition sequences that accurately reflect the described sequence.

\noindent{\bfseries Conditional splicing.}
The~\textit{timestamp stitcher} can add some control information, such as text or frame number, when splicing two short motion sequences,as shown in the figure~\ref{fig:app3}. For instance, the inputs to the module might be two independent motion sequences accompanied by the text descriptor "run". This text information is then embedded during the generation of the transition frame, guiding the transition sequence towards the "run" style.

Furthermore, The length of the transition sequence can also be controlled. Same as text control, users simply specify the desired frame number during input. The \textit{timestamp stitcher} then produces a transition sequence of the requisite length based on these frame number. Additionally, users can also use text and frame number control at the same time.

\noindent{\bfseries Unconditional splicing.}
The module is also capable of generating transition sequences unconditionally. Users simply need to input two independent human motion sequences with consistent human body representations into the model, and the model will seamlessly splice the two sequences into a complete sequence with arbitrarily length and style. Each output is unique, ensuring diversity in the results.

Our~\textit{timestamp stitcher} demonstrates the ability to generate transition sequences between two short motion sequences. It is adept at handling any  motion sequence and producing transitions that are both natural and smooth. Whether operating under conditional or unconditional settings, the~\textit{timestamp stitcher} is good at generating naturally fluid transition frames. This capability ensures the generation of stable, high-quality motion sequences, thereby enhancing the overall quality of infinite motion generation.
\begin{figure*}
\centering
\includegraphics[width=\textwidth]{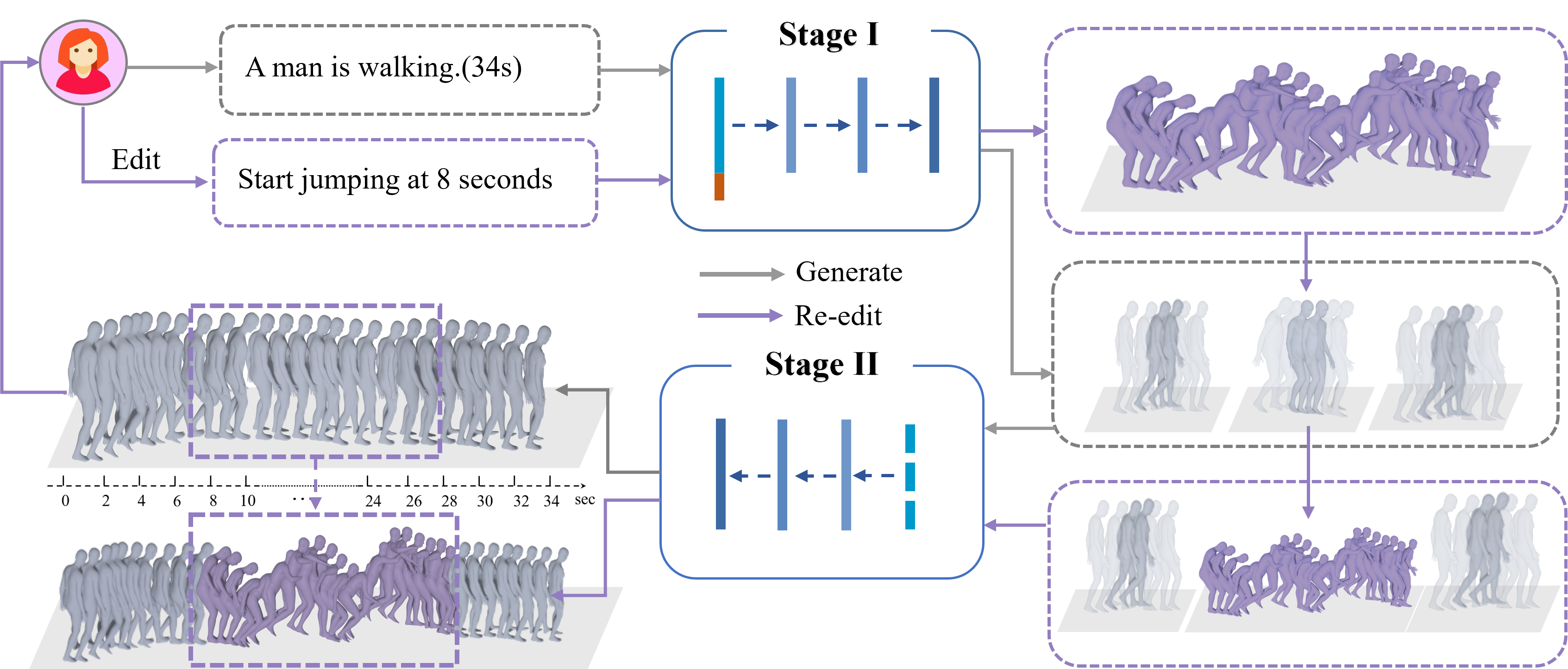}
\caption{The flow of editing motion segment: Initially, the user inputs text description, prompting Stage One, where the model outputs several short motion sequences. These short motion sequences serve as inputs for Stage Two, during which the~\textit{timestamp stitcher} stitches these short motion sequences to produce a complete, extended motion sequence. This sequence is reviewed by the user who may choose to edit it by re-inputting text, generating a new short sequence in Stage One. This new sequence replaces an original short sequence in Stage Two, along with the remaining sequences from the first iteration, resulting in a newly formed long sequence. The gray line in the diagram illustrates the user's initial generation, while the purple line indicates the user’s subsequent re-editing of the sequence.}
\label{fig:app2}
\end{figure*}
\begin{figure*}
\centering
\includegraphics[width=\textwidth]{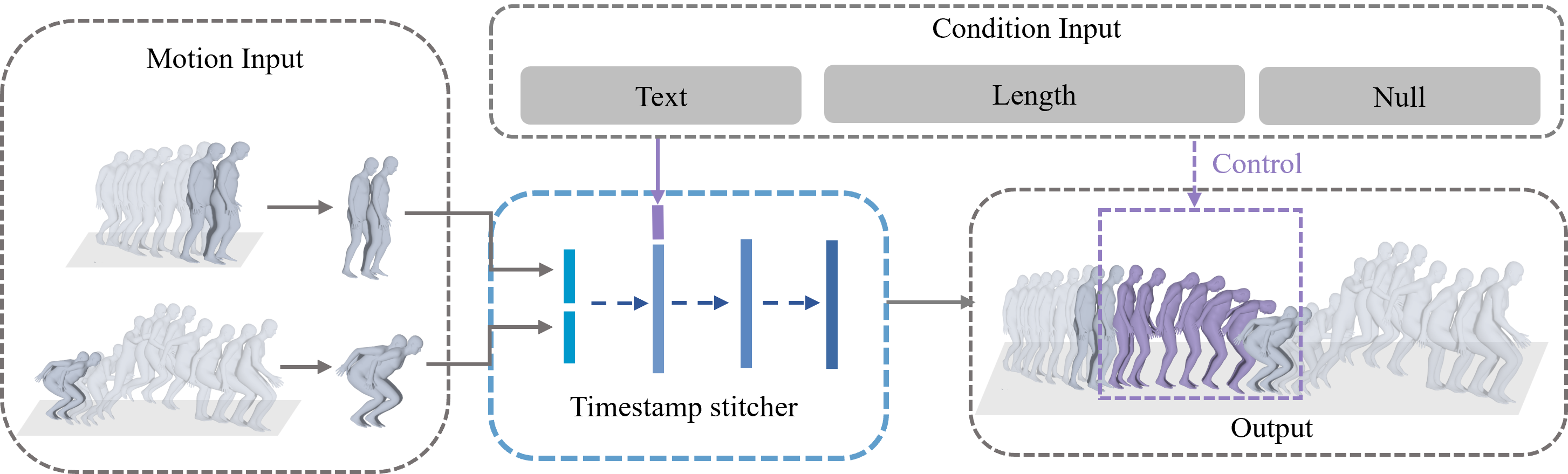}
\caption{The flow of splicing: Input any two motion sequences, ensuring that the human body representations in both segments are consistent. Extract the last 5 frames of the previous motion sequence and the first 5 frames of the next motion sequence, and input them into the~\textit{timestamp stitcher} for splicing. During the splicing process, text or length information can also be added to control the stitching style and length of the splicing sequence.}
\label{fig:app3}
\end{figure*}
\section{conclusion}
In this paper, we introduce HumanML3D-Extend dataset, an enhanced dataset designed for the generation of long human motion sequences driven by detailed textual descriptions.
HumanML3D-Extend dataset encompasses not only super-long motion data but also precise timestamps and comprehensive text annotations.
We conduct a multifaceted analysis of the HumanML3D-Extend dataset to demonstrate the usability of the dataset.
Building on this dataset, we propose Infinite Motion, a novel model capable of processing text descriptions of arbitrary length to generate limitless human motion sequences.
Different from traditional generative models, Infinite Motion is designed not only to generate complete sequences but also to edit individual segments, enhancing its flexibility and customizability. 
Additionally, we present three applications of our work that demonstrate the potential of our method in various settings, proving the effectiveness and versatility of the infinite motion model.
Extensive experimental evidence has confirmed that this dataset can be established as a reliable benchmark for generating long motion sequences, demonstrating the stability and efficiency of the novel splicing method we have developed in conjunction with this dataset.

\noindent{\bfseries Limitations.} Despite the advancements introduced by the HumanML3D-Extend dataset and the Infinite Motion model, several limitations remain. Firstly, the complexity and size of the dataset may pose challenges for training, requiring significant computational resources and potentially limiting accessibility for researchers with less powerful hardware. Furthermore, the seamless stitching of disparate sequences relies heavily on the consistency of human body representations, which cannot be guaranteed across different data sources. Lastly, the evaluation of the generated sequences' coherence and fluidity is still largely subjective, and developing more objective and quantifiable metrics for this purpose remains an open challenge. These limitations suggest avenues for future research, including optimizing computational efficiency, enhancing robustness to input variability, and developing better evaluation metrics.

\bibliographystyle{IEEEtran}

\end{document}